\documentclass[conference]{IEEEtran}
\IEEEoverridecommandlockouts
\usepackage[T1]{fontenc}
\usepackage{cite}
\usepackage{amsmath,amssymb,amsfonts}
\usepackage{stfloats}
\usepackage{graphicx}
\usepackage{textcomp}
\usepackage{xcolor}
\usepackage{booktabs,array}
\usepackage{url}
\usepackage[hidelinks]{hyperref}
\def\BibTeX{{\rm B\kern-.05em{\sc i\kern-.025em b}\kern-.08em
    T\kern-.1667em\lower.7ex\hbox{E}\kern-.125emX}}
\makeatletter
\newcommand{\linebreakand}{%
  \end{@IEEEauthorhalign}
  \hfill\mbox{}\par
  \mbox{}\hfill\begin{@IEEEauthorhalign}
}
\makeatother
\graphicspath{{figures/}}
\newcommand{\fbar}{\bar{f}}
\newcommand{\Bbar}{\bar{B}}

\newcommand{\MapeA}{1.89}
\newcommand{\TermsA}{10}
\newcommand{\RtwoA}{0.9998}

\newcommand{\MapeB}{1.01}
\newcommand{\TermsB}{10}
\newcommand{\RtwoB}{0.9995}

\newcommand{\MapeC}{1.56}
\newcommand{\TermsC}{7}
\newcommand{\RtwoC}{0.9997}

\begin{document}
\bstctlcite{BSTcontrol}
\title{Discovering Explicit Magnetic Core-Loss Equations via Symbolic Sparse Identification 
\thanks{\footnotesize This work has been accepted by the IEEE for publication. Copyright may be transferred without notice,
after which this version may no longer be accessible.
Source code available at: \url{https://github.com/Aaron-H-Wang/LSSI/tree/main/SSI}.}
}
\author{
\IEEEauthorblockN{Haoyu Wang}
\IEEEauthorblockA{\textit{Dept. Electrical and Computer Eng.} \\
\textit{The University of Texas at Austin}\\
wanghaoyu@utexas.edu}
\and
\IEEEauthorblockN{Jialin Zheng}
\IEEEauthorblockA{\textit{Dept. Electrical and Computer Eng.} \\
\textit{Princeton University}\\
jz8197@princeton.edu}
\and
\IEEEauthorblockN{Ziyang Xu}
\IEEEauthorblockA{\textit{Dept. Electrical and Computer Eng.} \\
\textit{The University of Texas at Austin}\\
ziyangxu@utexas.edu}
\linebreakand
\IEEEauthorblockN{Yihao Wu}
\IEEEauthorblockA{\textit{Dept. Electrical and Computer Eng.} \\
\textit{The University of Texas at Austin}\\
yw25243@utexas.edu}
\and
\IEEEauthorblockN{Alex Hanson}
\IEEEauthorblockA{\textit{Dept. Electrical and Computer Eng.} \\
\textit{The University of Texas at Austin}\\
ajhanson@utexas.edu}
}
\maketitle
\begin{abstract}
Compact core loss equations are essential tools in magnetic component design, but existing approaches trade accuracy against interpretability. Empirical models such as the Steinmetz Equation (SE) yield compact expressions of limited accuracy, while machine learning methods achieve high accuracy as black-box predictors. This paper bridges the gap with a Symbolic Sparse Identification (SSI) framework, which casts core loss modeling as a data-driven symbolic regression problem and recovers unknown equations directly from measured data, without presuming a functional form. First, a comprehensively expanded library of symbolic candidate terms far richer than the SE is constructed. Second, the governing terms are isolated from this library by sparse regression. Third, the coefficients of the retained terms are refined with an AdamW optimizer to maximize accuracy and form an analytical equation.  Experiments on Fair-Rite 78, 79, and 95 yield test mean absolute percentage errors of \MapeA\%, \MapeB\%, and \MapeC\%, respectively, with only \TermsA, \TermsB, and \TermsC\ active terms. Comparisons with empirical, numerical, and symbolic models quantify the resulting balance of accuracy and compactness. SSI thus provides an explicit and physically guided approach for high-frequency magnetic characterization and design.
\end{abstract}

\section{Introduction}
The need for higher power density in power electronics demands increasing operating frequencies, making magnetic core loss an important efficiency and thermal constraint \cite{7370781, 9893537, 10242261, 10132875, 10319663, 10362302, 9628583, 10194423, 11358392, 11366028}. It is therefore essential to have accurate loss prediction for material selection and magnetic design. An explicit equation as a predictor is exceptionally useful because it presents scaling relationships and can be directly integrated into optimization.

For more than a century, magnetic core loss has been characterized by the Steinmetz Equation (SE) and its variants \cite{1457110,5955126}. Under sinusoidal excitation at a fixed temperature, the SE expresses volumetric core loss as $P_v=kf^{\alpha}B^{\beta}$, where $f$ is the operating frequency and $B$ is the peak flux density. However, it also has the limitation that a three-parameter power law cannot precisely represent the nonlinear behavior of magnetic materials in a wide operating range, leading to structural errors and poor accuracy. Modified and generalized Steinmetz equations extend core-loss prediction to nonsinusoidal excitation \cite{Reinert2001MSE,Venkatachalam2002IGSE}, but prescribed analytical structures are retained and structural errors might still exist.

Recent machine learning (ML) approaches have largely improved the accuracy by capturing nonlinear magnetic characteristics from measured data, at the expense of relinquishing the explicit equation rather than refining it \cite{10169101}. Random forest \cite{10509368} and feedforward neural networks \cite{10502151} reproduce measured core loss with errors of a few percent by absorbing the loss nonlinearities into a learned mapping. However, they yield a numerical predictor rather than an analytical expression, and improved accuracy does not restore the scaling information that the SE conveys explicitly, which is insufficient for practical magnetic design. 

ML-based symbolic regression addresses this issue by identifying analytical expressions directly from data. Specifically, the state-of-the-art Sparse Identification of Nonlinear Dynamics (SINDy) selects functions from a predefined library \cite{Brunton2016SINDy}, whose principle can also be applied to the static loss surface. However, conventional polynomial sparse regression with arbitrary coefficient signs cannot represent nonnegative amplitudes of physical loss mechanisms. 

Following the path, this paper proposes a Symbolic Sparse Identification (SSI) framework to solve the core-loss equation discovery problem while maintaining physical interpretability. Specifically, SSI formulates the core-loss modeling as a symbolic sparse regression problem and constructs a fixed-grid physics-guided candidate library based on magnetic-loss behaviors. Dominant candidates are isolated by a hybrid objective, AdamW updates, and a pruning mechanism. A compact equation with nonnegative coefficients is ultimately identified from measurements. SSI therefore provides an explicit and physically interpretable approach for high-frequency magnetic characterization and design.

\section{Symbolic Sparse Identification Framework}

\begin{figure*}[t]
\setlength{\abovecaptionskip}{3pt}
\centering
\includegraphics[width=\textwidth]{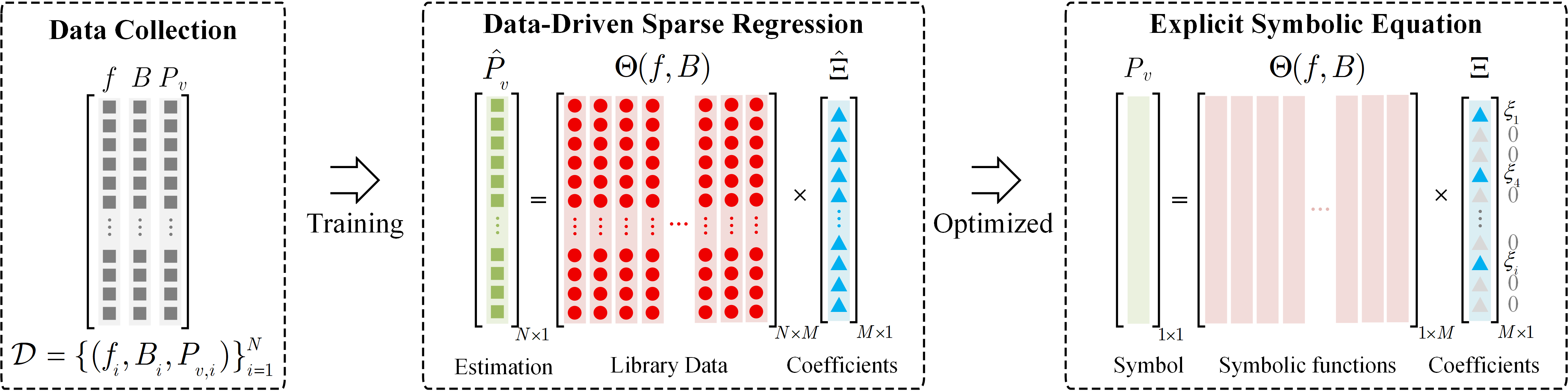}
\caption{The proposed SSI framework, including data collection, sparse identification over a fixed candidate library, and explicit equation discovery.}
\label{fig:framework}
\end{figure*}

The proposed SSI framework is shown in Fig.~\ref{fig:framework}. Specifically, a fixed symbolic library is first evaluated using normalized experimental data and its candidate coefficients are then optimized. Weak candidates will be pruned and the retained coefficients are refined to obtain an explicit core-loss equation. The details are introduced below.

\subsection{Problem Formulation of Core-Loss Modeling}
In macroscopic magnetic characterization, measured core loss can be regarded as an aggregate response with nonlinear dependencies. Classical core-loss descriptions present hysteresis, eddy-current, and excess contributions, while high-frequency behavior introduces further curvature \cite{Bertotti1988,1041954}. Motivated by the core-loss theory, this paper represents the loss surface as a sparse combination of nonlinear functions.

Specifically, let $\mathcal{D}=\{(f_i,B_i,P_{v,i})\}_{i=1}^{N}$ denote the measurements under sinusoidal excitation at a fixed temperature. Define the training-set means as $f_n$, $B_n$, and $P_{v,n}$. The dimensionless variables can therefore be expressed as:
\begin{equation}
\bar{f}=\frac{f}{f_n},\qquad
\bar{B}=\frac{B}{B_n},\qquad
\bar{P}_v=\frac{P_v}{P_{v,n}}.
\label{eq:normalization}
\end{equation}

The normalized equation-discovery problem is expressed as:
\begin{equation}
\bar{P}_v=\boldsymbol{\Theta}(\bar{f},\bar{B})\cdot\boldsymbol{\Xi},
\qquad\boldsymbol{\Xi}\succeq\boldsymbol{0}
\label{eq:ssi_form}
\end{equation}
where $\boldsymbol{\Theta}=[\theta_1,\theta_2,\ldots,\theta_M]$ is a row vector of candidate functions and $\boldsymbol{\Xi}=[\xi_1,\xi_2,\ldots,\xi_M]^T$ is a coefficient vector. 

The purpose of sparse identification is to optimize $\boldsymbol{\Xi}$ and retain only a few nonzero entries of it, thus selecting the dominant functions that finally form the explicit equation. With the retained set $\mathcal{S}$, the dimensional core-loss prediction is then recovered as:

\begin{equation}
\hat{P}_v(f,B)=P_{v,n}\hat{\bar{P}}_v=P_{v,n}\sum_{j\in\mathcal{S}}
\xi_j\theta_j\!\left(\frac{f}{f_n},\frac{B}{B_n}\right).
\label{eq:dimensional}
\end{equation}

\subsection{Physics-Guided Candidate Library}

The symbolic candidate library determines which nonlinear relationships can be identified. Therefore, a good library is important to the SSI framework. To represent a wide range of behaviors, SSI combines the following groups of functions:
\begin{equation}
\begin{aligned}
\boldsymbol{\Theta}(\bar{f},\bar{B})=\Big[&
\underbrace{\bar{f}^{\alpha}\bar{B}^{\beta}}_{\text{power laws}},\; 
\underbrace{\frac{\bar{f}^{\alpha}\bar{B}^{\beta}}
{1+(\bar{f}/\bar{f}_c)^2}}_{\text{high-frequency roll-off}},\\
&
\underbrace{\bar{f}^{\alpha}\bar{B}^{\beta}\ln\bar{B},\;
\bar{f}^{\alpha}\bar{B}^{\beta}\ln\bar{f}}_{\text{exponent corrections}},\\
&
\underbrace{\bar{f}\bar{B}^2(e^{\bar{B}/\bar{B}_c}-1)}_{\text{saturation-related}},\;
\underbrace{\bar{f}\bar{B},\bar{f},\bar{B},1}_{\text{auxiliaries}}\Big].
\end{aligned}
\label{eq:library}
\end{equation}

\subsubsection{Power-Law Terms}
The power-law functions with fractional frequency and flux-density exponents are used to generalize the SE, including hysteresis-like and eddy-current-like nonlinear dependencies. A separate excess-loss-like term $\bar{f}^{1.5}\bar{B}^{1.5}$ is also included, providing 73 candidates in total.

\subsubsection{High-Frequency Roll-Off Terms}
A frequency-dependent attenuation $1/[1+(\bar{f}/\bar{f}_c)^2]$ with a dimensionless cutoff frequency $\bar{f}_c$ is introduced into high-frequency roll-off functions. These 18 terms can capture effective high-frequency curvature that cannot be represented by constant exponents alone. 

\subsubsection{Logarithmic Exponent Corrections}
An exponent shift can be approximated by
\begin{equation}
\begin{aligned}
&\bar{f}^{\alpha+\Delta\alpha}
=\bar{f}^{\alpha}e^{\Delta\alpha\ln\bar{f}}
\approx\bar{f}^{\alpha}+\Delta\alpha\bar{f}^{\alpha}\ln\bar{f}.\\
&\bar{B}^{\beta+\Delta\beta}
=\bar{B}^{\beta}e^{\Delta\beta\ln\bar{B}}
\approx\bar{B}^{\beta}+\Delta\beta\bar{B}^{\beta}\ln\bar{B}.
\end{aligned}
\label{eq:drift}
\end{equation}

Therefore, $\bar{f}^{\alpha}\bar{B}^{\beta}\ln\bar{f}$ and $\bar{f}^{\alpha}\bar{B}^{\beta}\ln\bar{B}$ are first-order corrections associated with exponent changes. These six functions supplement the power-law grid.

\subsubsection{Saturation-Related and Auxiliary Functions}
Two exponential candidates represent a rapid loss increase at high flux density, and four low-order terms allow cross-coupling and offset corrections. 

To sum up, these blocks give $M=103$ functions in total. Note that all internal exponents and other parameters are fixed during identification, while the coefficients are optimized with non-negativity to represent positive dissipation. Table~\ref{tab:library} shows the fixed-grid candidate library used in the SSI framework. The fixed-grid sets are defined as:

% Nonnegative $\xi_j$ constrain the amplitude of each prescribed candidate. The logarithmic functions themselves can change sign and serve as curvature corrections; they are not interpreted as independently positive dissipative components.

\begin{equation}
\begin{aligned}
\mathcal{A}&=\{1,1.2,1.4,1.5,1.6,1.8,2,2.2\}; \\
\mathcal{B}&=\{1.6,1.8,2,2.2,2.4,2.5,2.6,2.8,3\}; \\
\mathcal{P}_{h}&=\{(1,2),(1.5,2.4),(2,2)\};\\
\mathcal{P}_{\ell}&=\{(1,2),(1.4,2.4),(1.8,2.4)\};\\
\mathcal{F}_c&=\{0.5,0.75,1,1.5,2,3\}.    
\end{aligned}
\end{equation}

\begin{table}[t]
\centering
\caption{Physics-Guided Fixed Candidate Library for SSI}
\label{tab:library}
\footnotesize
\begin{tabular}{cllc}
\toprule
Indices & Candidate $\theta_j(\bar{f},\bar{B})$ & Fixed parameter choices & Count \\
\midrule
1--72 & $\bar{f}^{\alpha}\bar{B}^{\beta}$ & $\alpha\in\mathcal{A}$, $\beta\in\mathcal{B}$ & 72 \\
73 & $\bar{f}^{1.5}\bar{B}^{1.5}$ & Excess-loss-like & 1 \\
74--91 & $\bar{f}^{\alpha}\bar{B}^{\beta}/[1+(\bar{f}/\bar{f}_c)^2]$ & $(\alpha,\beta)\in\mathcal{P}_{h}$, $\bar{f}_c\in\mathcal{F}_c$ & 18 \\
92--97 & $\bar{f}^{\alpha}\bar{B}^{\beta}\ln\bar{B}$, $\bar{f}^{\alpha}\bar{B}^{\beta}\ln\bar{f}$ & $(\alpha,\beta)\in\mathcal{P}_{\ell}$ & 6 \\
98--99 & $\bar{f}\bar{B}^{2}(e^{\bar{B}/\bar{B}_c}-1)$ & $\bar{B}_c\in\{2,4\}$ & 2 \\
100--103 & $\bar{f}\bar{B}$, $\bar{f}$, $\bar{B}$, $1$ & Low-order corrections & 4 \\
\bottomrule
\end{tabular}
\par\smallskip
\begin{minipage}{0.48\textwidth}\footnotesize
*Power-law exponents follow discrete $\alpha$ and $\beta$; roll-off parameters follow ascending $\bar{f}_c$ and the pair order $\mathcal{P}_{h}$; and logarithmic exponents follow the pair order $\mathcal{P}_{\ell}$.
\end{minipage}
\end{table}

\subsection{Hybrid Objective and Sparse Identification}

The numerical magnitudes of the library columns may significantly differ after input normalization. Therefore, root-mean-square scales are defined in the training set:
\begin{equation}
s_j=\sqrt{\frac{1}{N}\sum_{i=1}^{N}\theta_j^2(\bar{f}_i,\bar{B}_i)}+\epsilon,
\label{eq:column_scale}
\end{equation}
where $\epsilon=10^{-8}$ is a numerical stabilizer. With $\tilde{\theta}_j=\theta_j/s_j$ and $\tilde{\xi}_j=s_j\xi_j$, the prediction remains unchanged as:
\begin{equation}
\hat{\bar{P}}_v=\boldsymbol{\Theta}\cdot\boldsymbol{\Xi}
=\tilde{\boldsymbol{\Theta}}\cdot\tilde{\boldsymbol{\Xi}}.
\label{eq:scaled_prediction}
\end{equation}

% Thus, a common threshold acting on $\tilde{\boldsymbol{\Xi}}$ can be applied across function types. 
% The final reported coefficients are recovered as $\xi_j=\tilde{\xi}_j/s_j$.

Core loss also spans several orders of magnitude. An absolute-error objective may consequently be dominated by high-flux-density samples. Therefore, SSI introduces logarithmic and percentage errors into the training objective:
\begin{equation}
\begin{aligned}
\mathcal{L}(\tilde{\boldsymbol{\Xi}})={}
\underbrace{\frac{w_1}{N}\sum_{i=1}^{N}
\left(\ln\hat{\bar{P}}_{v,i}-\ln\bar{P}_{v,i}\right)^2}
_{\text{logarithmic term}}+\underbrace{\frac{w_2}{N}\sum_{i=1}^{N}
\left|\frac{\hat{\bar{P}}_{v,i}-\bar{P}_{v,i}}{\bar{P}_{v,i}}\right|}
_{\text{MAPE term}},
\end{aligned}
\label{eq:objective}
\end{equation}
where the logarithmic term shapes the identification across the loss range, while the Mean Absolute Percentage Error (MAPE) term directly targets relative accuracy. Equal weights $w_1=w_2=0.5$ are used. 
% In implementation, predicted loss is clamped below by $\epsilon=10^{-8}$ before the logarithm; $\epsilon$ is added to targets inside logarithms and denominators. These stabilizers are omitted from~\eqref{eq:objective} for clarity.

The objective is minimized using AdamW, which combines adaptive gradient
moments with decoupled weight decay
\cite{Kingma2015Adam,Loshchilov2019AdamW}.
At iteration $k\geq1$, the gradient with respect to the optimized
coefficients is
\begin{equation}
\boldsymbol{g}^{(k)}
=\nabla_{\tilde{\boldsymbol{\Xi}}}
\mathcal{L}\!\left(\tilde{\boldsymbol{\Xi}}^{(k-1)}\right).
\label{eq:adamw_gradient}
\end{equation}

The first and second moments are updated as
\begin{equation}
\begin{aligned}
\boldsymbol{m}^{(k)}
&=\sigma_1\boldsymbol{m}^{(k-1)}
 +(1-\sigma_1)\boldsymbol{g}^{(k)},\\
\boldsymbol{v}^{(k)}
&=\sigma_2\boldsymbol{v}^{(k-1)}
 +(1-\sigma_2)
 \boldsymbol{g}^{(k)}\odot\boldsymbol{g}^{(k)}
\end{aligned}
\label{eq:adamw_moments}
\end{equation}
where $\sigma_1=0.9$ and $\sigma_2=0.999$ are moment-decay factors,
and $\odot$ denotes elementwise multiplication.
Starting from
$\boldsymbol{m}^{(0)}=\boldsymbol{v}^{(0)}=\boldsymbol{0}$,
the initialization bias is corrected by
\begin{equation}
\hat{\boldsymbol{m}}^{(k)}
=\frac{\boldsymbol{m}^{(k)}}{1-\sigma_1^{k}},
\qquad
\hat{\boldsymbol{v}}^{(k)}
=\frac{\boldsymbol{v}^{(k)}}{1-\sigma_2^{k}}.
\label{eq:adamw_bias_correction}
\end{equation}

The coefficient update and nonnegative projection are:
\begin{equation}
\begin{aligned}
\boldsymbol{z}^{(k)}
&=(1-\eta_k\gamma)\tilde{\boldsymbol{\Xi}}^{(k-1)}
-\eta_k
\frac{\hat{\boldsymbol{m}}^{(k)}}
{\sqrt{\hat{\boldsymbol{v}}^{(k)}}+\epsilon},\\
\tilde{\boldsymbol{\Xi}}^{(k)}
&=\max\!\left[\boldsymbol{z}^{(k)},\boldsymbol{0}\right]
\end{aligned}
\label{eq:adamw}
\end{equation}
where $\eta_k$ and $\gamma$ are the learning rate and
weight-decay rate, respectively. Note that $\gamma$ acts directly on the coefficients rather than the gradient moments.

A binary support mask initialized as
$\boldsymbol{d}^{(0)}=\boldsymbol{1}$ is then introduced for pruning.
At pruning checkpoints, a gradually increasing threshold $\tau^{(k)}$ removes weak candidates:

\begin{equation}
\begin{aligned}
d_j^{(k)}
&=d_j^{(k-1)}
\mathbb{I}\!\left[
\tilde{\xi}_j^{(k)}\geq\tau^{(k)}
\right],\\
\tilde{\boldsymbol{\Xi}}^{(k)}
&\leftarrow
\tilde{\boldsymbol{\Xi}}^{(k)}\odot\boldsymbol{d}^{(k)}
\end{aligned}
\label{eq:mask_update}
\end{equation}
where $\mathbb{I}[\cdot]$ is the indicator function that outputs 1 or 0.
Between checkpoints, the mask remains unchanged. Note that the pruning is irreversible. Sparsity is imposed by projection and threshold masking,
without an explicit $\ell_1$ penalty.

To sum up, coefficient discovery consists of three stages:
\begin{enumerate}
    \item a warm-up stage, where all candidates are
    available and coefficients establish their initial scales
    without weight decay;
    \item a discovery stage, where AdamW with weight decay is applied with the settings shown in Table~\ref{tab:hyper};
    \item a delayed pruning stage, where weak candidates are
    progressively removed using the increasing threshold.
    % \item a support-fixed refinement stage, where the retained
    % coefficients are refined with the selected support fixed.
\end{enumerate}

\begin{table}[t]
\centering
\caption{Hyperparameters for SSI Training}
\label{tab:hyper}
\footnotesize
\setlength{\tabcolsep}{3pt}
\begin{tabular}{lclc}
\toprule
Parameter & Value & Parameter & Value \\
\midrule
Library size $M$ & 103 & $w_1$, $w_2$ & 0.5, 0.5 \\
Warm-up updates & 1200 & Warm-up $\eta$ & 0.01 \\
Pruning onset & 30\% & Initial $\eta$ & 0.005 \\
Discovery updates & 30,001 & $\tau_{\max}$ & 0.1 \\
Weight decay $\gamma$ & 0.05 & $\epsilon$ & $10^{-8}$ \\
\bottomrule
\end{tabular}
\end{table}

\section{Experimental Validation}
\subsection{Experimental Data and Implementation}
The proposed SSI framework is validated on high-fidelity measurements of Fair-Rite 78, 79, and 95 ferrites, denoted FR78, FR79, and FR95, respectively. The datasets were collected under sinusoidal excitation at 25$^{\circ}$C using the automated resonant measurement system in \cite{10977470, 11121265, Wang2026Measurement}, as shown in Fig.~\ref{fig:domains}. The data specifications are listed in Table~\ref{tab:data}. All modeling methods use the same 80/20 training/test split, and SSI is implemented in PyTorch with the shared hyperparameters in Table~\ref{tab:hyper}. Metrics are then evaluated after dimensional loss is recovered using~\eqref{eq:dimensional}. For $N_t$ test samples, the MAPE, coefficient of determination, and percentage residuals are:

\begin{figure*}[t]
\setlength{\abovecaptionskip}{3pt}
\centering
\includegraphics[width=\textwidth]{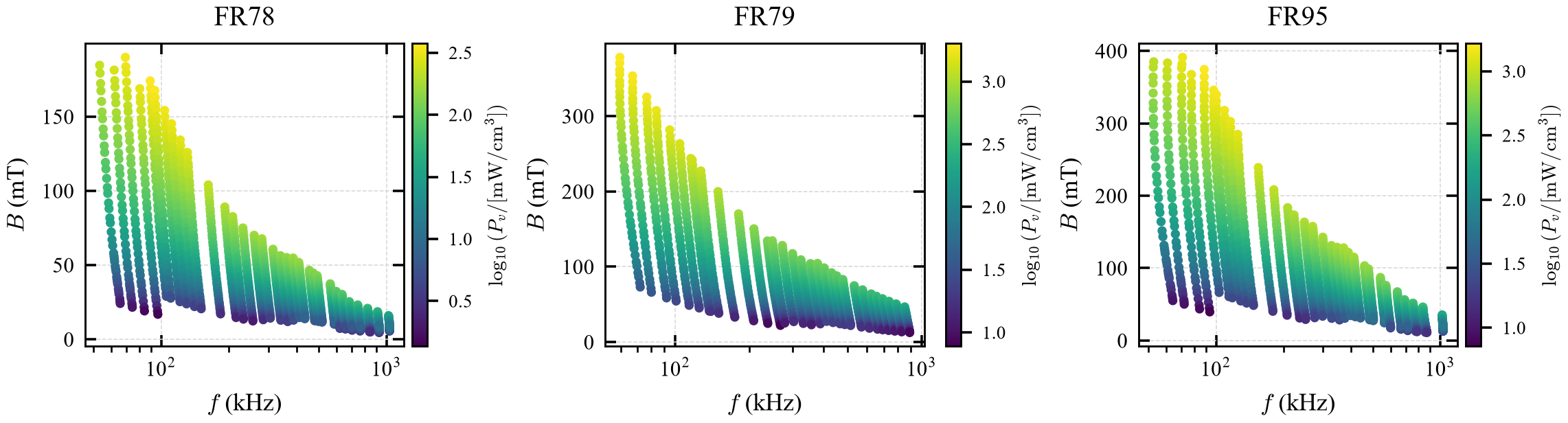}
\caption{Measured operating points of FR78, FR79, and FR95.}
\label{fig:domains}
\end{figure*}

\begin{table}[t]
\centering
\caption{Experimental Data and Normalization}
\label{tab:data}
\scriptsize
\setlength{\tabcolsep}{3pt}
\begin{tabular}{lccc}
\toprule
Quantity & FR78 & FR79 & FR95 \\
\midrule
Train/test samples & 1056/264 & 992/248 & 1056/264 \\
$f$ (kHz) & 53.4--1030 & 59.5--893 & 52.4--1029 \\
$B$ (mT) & 4.52--189.92 & 12.53--378.00 & 10.97--391.09 \\
$P_v$ (mW/cm$^3$) & 1.37--374.24 & 7.76--2004.31 & 7.16--1643.80 \\
$f_{n}$ (kHz) & 382.7853 & 360.3162 & 372.5723 \\
$B_{n}$ (mT) & 44.7138 & 88.4504 & 102.5971 \\
$P_{v,n}$ (mW/cm$^3$) & 59.7972 & 288.6610 & 303.9457 \\
\bottomrule
\end{tabular}
\end{table}
\vspace{-5pt}

\begin{align}
\mathrm{MAPE}&=\frac{100\%}{N_t}\sum_{i=1}^{N_t}
\left|\frac{\hat{P}_{v,i}-P_{v,i}}{P_{v,i}}\right|.\\
R^2&=1-\frac{\sum_{i=1}^{N_t}(\hat{P}_{v,i}-P_{v,i})^2}
{\sum_{i=1}^{N_t}\left(P_{v,i}-\frac{1}{N_t}\sum_{\ell=1}^{N_t}P_{v,\ell}\right)^2}.\\
r_i&=(\hat{P}_{v,i}-P_{v,i})/P_{v,i}\times 100\%.
\label{eq:metrics}
\end{align}

\subsection{Discovered Core-Loss Equations}
The identified equations retain \TermsA, \TermsB, and \TermsC\ terms for FR78, FR79, and FR95, respectively, as listed in Table~\ref{tab:coefficients}. Taking FR95 as an example, the seven-term equation can be written as follows:

\begin{equation}
\begin{aligned}
\hat{\bar{P}}_v={}&
\underbrace{\sum_{j\in\{74,77,91\}}
\frac{\xi_j\bar{f}^{\alpha_j}\bar{B}^2}
{1+(\bar{f}/\bar{f}_{c,j})^2}}_{\text{high-frequency roll-off}}+\underbrace{\xi_{73}\bar{f}^{1.5}\bar{B}^{1.5}}
_{\text{excess-loss-like}}\\
&+\underbrace{\xi_{92}\bar{f}\bar{B}^2\ln\bar{B}
+\xi_{94}\bar{f}^{1.4}\bar{B}^{2.4}\ln\bar{B}}
_{\text{flux-exponent corrections}}\\
&+\underbrace{\xi_{95}\bar{f}^{1.4}\bar{B}^{2.4}\ln\bar{f}}
_{\text{frequency-exponent correction}}.
\end{aligned}
\label{eq:discovered}
\end{equation}

\begin{table}[t]
\centering
\caption{Identified Coefficients of the Discovered Core-Loss Equations}
\label{tab:coefficients}
\footnotesize
\setlength{\tabcolsep}{9pt}
\begin{tabular}{lcccc}
\toprule
$j$ & $\theta_j(\bar{f},\bar{B})$ & $\xi_j$ (FR78) & $\xi_j$ (FR79) & $\xi_j$ (FR95) \\
\midrule
3 & $\fbar\Bbar^{2}$ & -- & 0.1853 & -- \\
4 & $\fbar\Bbar^{2.2}$ & -- & 0.1749 & -- \\
13 & $\fbar^{1.2}\Bbar^{2.2}$ & -- & 0.1648 & -- \\
41 & $\fbar^{1.6}\Bbar^{2.4}$ & -- & 0.2133 & -- \\
46 & $\fbar^{1.8}\Bbar^{1.6}$ & 0.4646 & -- & -- \\
51 & $\fbar^{1.8}\Bbar^{2.5}$ & -- & 0.2733 & -- \\
73 & $\fbar^{1.5}\Bbar^{1.5}$ & -- & -- & 0.2983 \\
74 & $\tfrac{\fbar\Bbar^{2}}{1+(\fbar/0.5)^2}$ & 0.3116 & 0.2244 & 0.4820 \\
77 & $\tfrac{\fbar\Bbar^{2}}{1+(\fbar/0.75)^2}$ & 0.5790 & -- & 0.4733 \\
80 & $\tfrac{\fbar\Bbar^{2}}{1+(\fbar/1)^2}$ & 0.1351 & -- & -- \\
81 & $\tfrac{\fbar^{1.5}\Bbar^{2.4}}{1+(\fbar/1)^2}$ & 0.3213 & -- & -- \\
90 & $\tfrac{\fbar^{1.5}\Bbar^{2.4}}{1+(\fbar/3)^2}$ & -- & 0.2792 & -- \\
91 & $\tfrac{\fbar^{2}\Bbar^{2}}{1+(\fbar/3)^2}$ & 0.7575 & -- & 1.2731 \\
92 & $\fbar\Bbar^{2}\ln\Bbar$ & 0.4819 & 0.1907 & 0.4590 \\
94 & $\fbar^{1.4}\Bbar^{2.4}\ln\Bbar$ & -- & -- & 0.3982 \\
95 & $\fbar^{1.4}\Bbar^{2.4}\ln\fbar$ & 0.2388 & -- & 0.4052 \\
96 & $\fbar^{1.8}\Bbar^{2.4}\ln\Bbar$ & 0.4927 & -- & -- \\
97 & $\fbar^{1.8}\Bbar^{2.4}\ln\fbar$ & 0.4279 & -- & -- \\
98 & $\fbar\Bbar^2[e^{(\Bbar/2)}-1]$ & -- & 0.0561 & -- \\
99 & $\fbar\Bbar^2[e^{(\Bbar/4)}-1]$ & -- & 0.2393 & -- \\
\bottomrule
\end{tabular}
\par\smallskip
\begin{minipage}{0.48\textwidth}\footnotesize
*All omitted coefficients are zero.
\end{minipage}
\end{table}
\vspace{-5pt}

% Here $(\alpha_{74},\alpha_{77},\alpha_{91})=(1,1,2)$ and $(\bar{f}_{c,74},\bar{f}_{c,77},\bar{f}_{c,91})=(0.5,0.75,3)$ are fixed library values. The first group represents frequency-dependent curvature, the logarithmic groups correct the apparent exponents, and the last term supplies an excess-loss-like dependence.

The other two materials select different combinations from the same fixed-grid library. Specifically, FR78 retains several roll-off and logarithmic functions, whereas FR79 retains more unattenuated power laws and both exponential high-flux candidates. 
% Fig.~\ref{fig:families} groups their absolute contributions using~\eqref{eq:contribution}. 
The result shows how a common identification procedure accommodates different measured loss surfaces through material-dependent coefficients.

% \begin{figure}[t]
% \setlength{\abovecaptionskip}{3pt}
% \centering
% \includegraphics[width=\columnwidth]{candidate_families.png}
% \caption{Normalized absolute contributions of the retained candidate families. Contributions are calculated from $c_j$ in~\eqref{eq:contribution}.}
% \label{fig:families}
% \end{figure}

\subsection{Cross-Material Accuracy Validation}
\begin{figure*}[t]
\setlength{\abovecaptionskip}{3pt}
\centering
\includegraphics[width=\textwidth]{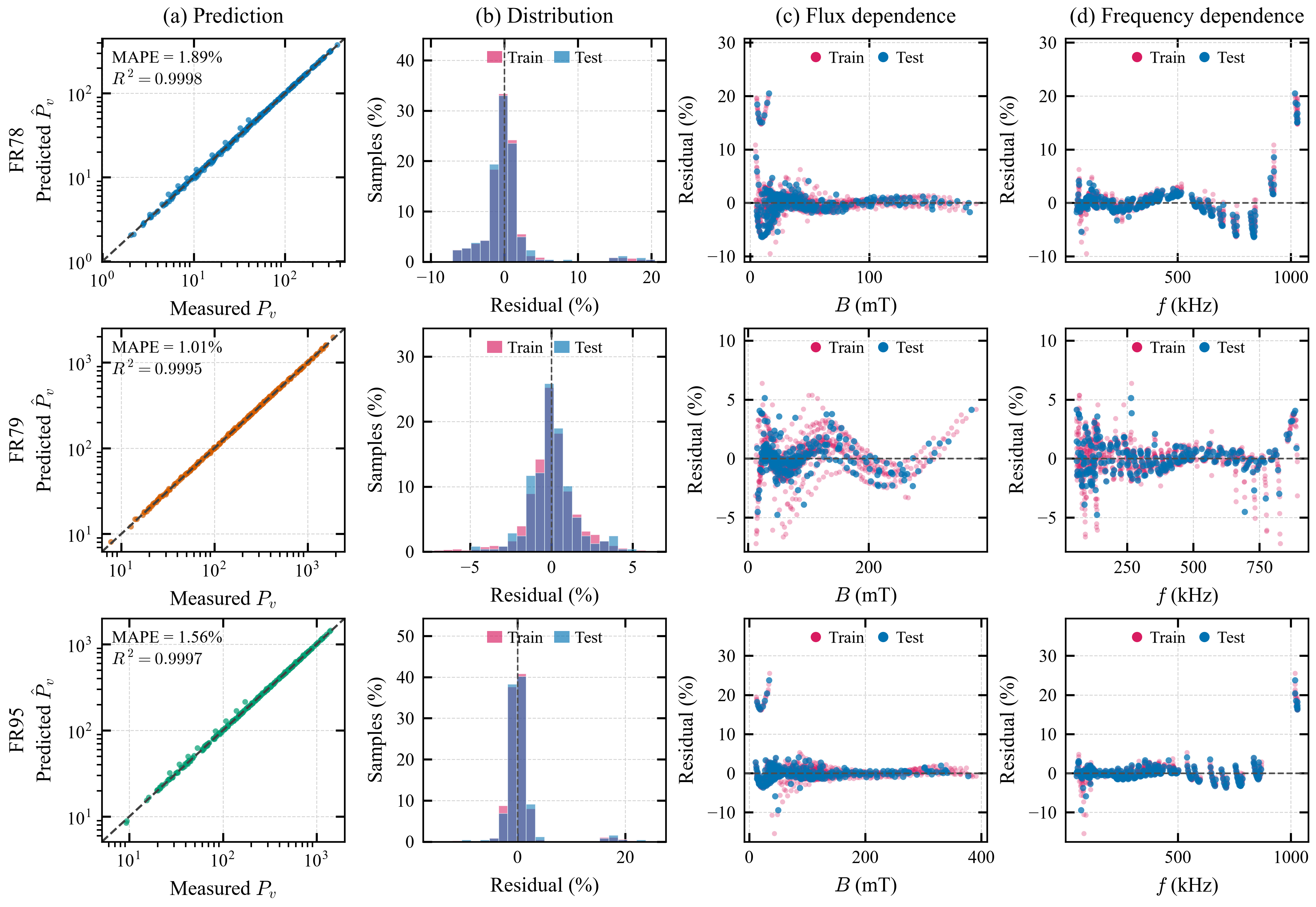}
\caption{Accuracy validation of SSI for FR78, FR79, and FR95. (a) Test predictions versus measurements. (b) Residual distributions. (c) Residuals versus flux density. (d) Residuals versus frequency.}
\label{fig:diagnostics}
\end{figure*}

The accuracy diagnostics for each material are shown in Fig.~\ref{fig:diagnostics}. Specifically, Fig.~\ref{fig:diagnostics}(a) shows that the predictions of test samples follow the measured loss across the logarithmic range. The test MAPE values are \MapeA\%, \MapeB\%, and \MapeC\% for FR78, FR79, and FR95, respectively, with corresponding $R^2$ values of \RtwoA, \RtwoB, and \RtwoC. The root-mean-square errors are 1.13, 7.52, and 5.61~mW/cm$^3$. All these results demonstrate that the discovered sparse equations capture the dominant nonlinear dependencies across different measured domains.
The residual distributions are shown in Fig.~\ref{fig:diagnostics}(b). Test residual means are $0.21\%$, $0.06\%$, and $0.46\%$, with standard deviations of $3.60\%$, $1.42\%$, and $3.54\%$. 
% FR79 has the narrowest distribution and the lowest relative error of the three SSI models.
Figs.~\ref{fig:diagnostics}(c) and (d) locate the remaining errors in the excitation domain. FR78 and FR95 show larger deviations in the high-frequency range with maximum test residuals of $20.47\%$ and $23.73\%$, respectively, compared with $5.13\%$ for FR79. Consequently, MAPE summarizes average accuracy and should be considered together with the residual distributions for magnetic designs. 
% Note that all final predictions are positive at the measured points.

\subsection{Training Evolution}
\begin{figure*}[t]
\setlength{\abovecaptionskip}{3pt}
\centering
\includegraphics[width=\textwidth]{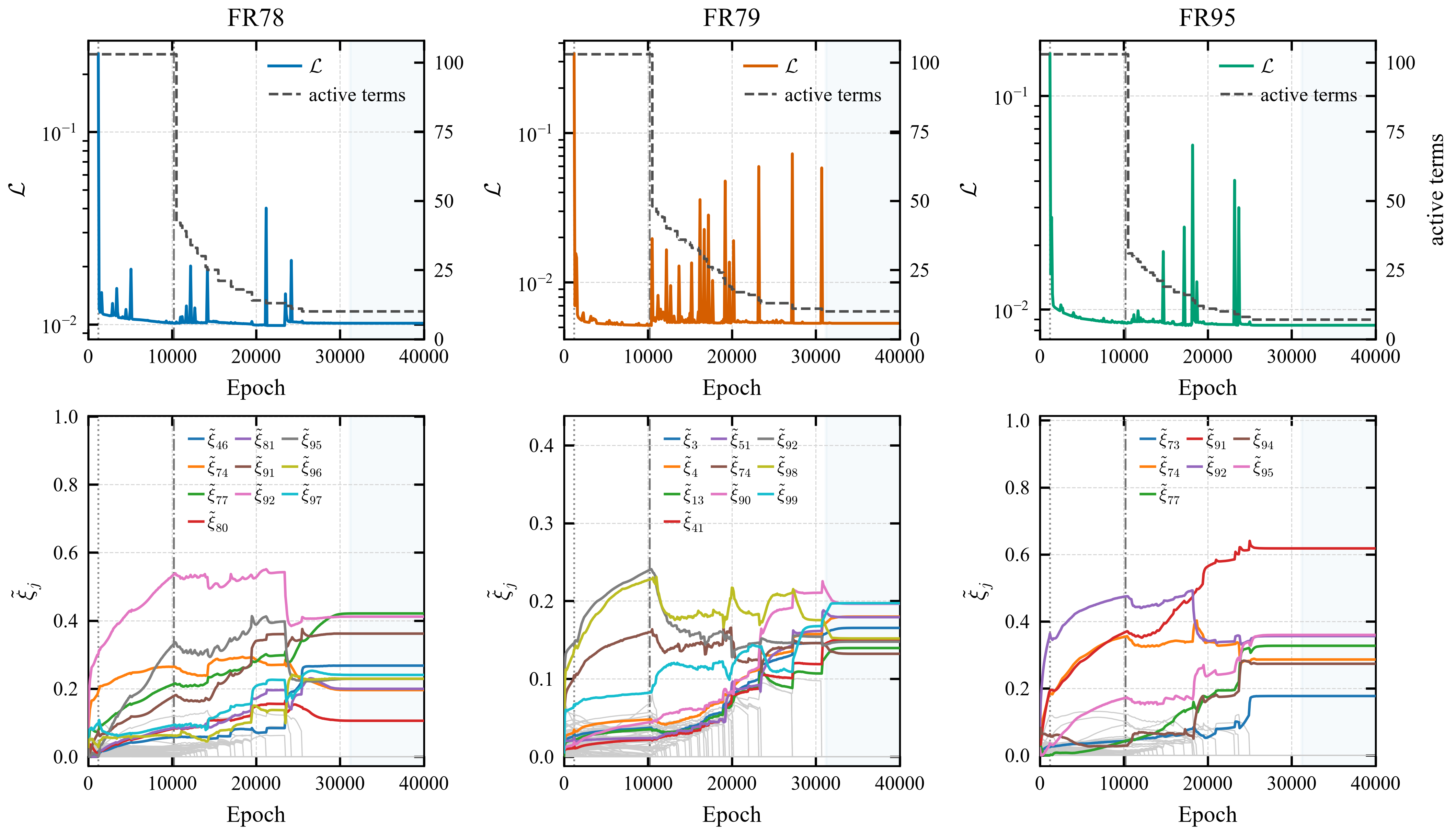}
\caption{Training evolution of SSI for FR78, FR79, and FR95. Top: hybrid objective $\mathcal{L}$ and number of active terms. Bottom: scaled coefficients $\tilde{\xi}_j$, with retained functions in color and other candidates in gray. 
% Dotted lines mark pruning onset in the top row and the end of warm-up in the bottom row. Final support-fixed refinement follows these trajectories.
}
\label{fig:training}
\end{figure*}

The training histories in Fig.~\ref{fig:training} show how the sparse equations are identified. It can be seen that the objective decreases rapidly during the initial updates. When delayed pruning begins, the active term count falls in discrete steps, while the surviving coefficients adjust according to the reduced library. Ramped thresholding progressively reduces the representation from 103 initial functions to at most ten retained terms, and the retained amplitudes stabilize during fixed-support continuation. The 1200 warm-up updates are followed by 30,001 discovery updates, with pruning starting after the first 30\% of discovery (approximately cumulative epoch 10,200). The saved runs then continue AdamW fitting on fixed support to cumulative epoch 60,000 before final refinement. Figure~\ref{fig:training} displays only the first 40,000 cumulative updates, including the initial part of this continuation. 

\subsection{Comparison With Other Modeling Methods}
The proposed SSI is compared with SE, RF, FNN, and a polynomial SINDy model on the same training and test samples. Specifically, the SE is fitted by least squares in logarithmic coordinates, the RF contains 200 trees, and the FNN uses two 64-unit ReLU hidden layers with 4,417 weights and biases. The SINDy baseline applies sequentially thresholded least squares to a polynomial library in standardized logarithmic coordinates $u_f$ and $u_B$: 
% Both numerical predictors are fitted to logarithmic loss using logarithmic frequency and flux-density inputs; FNN inputs are additionally standardized.

\begin{equation}
\log_{10}\hat{P}_v=\sum_{p+q\leq5}c_{pq}u_f^p u_B^q,
\qquad c_{pq}\in\mathbb{R}.
\label{eq:sindy}
\end{equation}

% There are 21 candidate monomials. Threshold selection uses every fifth outer-training point as an inner validation set and minimizes validation MAPE plus 0.005 times the term count. The selected thresholds are 0.5, 0.2, and 0.2 for FR78, FR79, and FR95. Coefficients are then refitted on all outer-training samples. This comparison concerns the specified unrestricted polynomial baseline.

\begin{table*}[t]
\centering
\caption{Comparison of Core-Loss Modeling Methods on the Same Test Sets}
\label{tab:comparison}
\footnotesize
\setlength{\tabcolsep}{5pt}
\begin{tabular}{lccccc}
\toprule
Method & \multicolumn{3}{c}{Test MAPE (\%)} & Model size: FR78 / FR79 / FR95 & Representation \\
 & FR78 & FR79 & FR95 & & \\
\midrule
SE & 12.76 & 2.45 & 10.65 & 3 / 3 / 3 & One power law \\
RF & 3.89 & 3.72 & 3.58 & 266732 / 250886 / 266732 & 200-tree numerical predictor \\
FNN & 0.40 & 0.70 & 0.49 & 4417 / 4417 / 4417 & Two 64-unit hidden layers \\
SINDy & 1.67 & 0.86 & 0.93 & 15 / 19 / 20 & Explicit signed polynomial in log coordinates \\
SSI & 1.89 & 1.01 & 1.56 & 10 / 10 / 7 & Nonnegative physics-guided equation \\
\bottomrule
\end{tabular}
\par\smallskip
% \begin{minipage}{0.98\textwidth}\footnotesize
% Fitted size denotes SE parameters, stored RF node quantities, FNN weights and biases, or active SINDy/SSI coefficients. These are method-specific size measures; fixed function definitions and normalization constants are additional model information.
% \end{minipage}
\end{table*}

Table~\ref{tab:comparison} summarizes the comparison of accuracy and model size. Specifically, the SE has test MAPEs of $12.76\%$, $2.45\%$, and $10.65\%$, indicating the limitation of one power law over the measured domains. The RF yields approximately $3.6$--$3.9\%$ error. The FNN achieves the lowest numerical errors, $0.40$--$0.70\%$, but its loss dependence is distributed across thousands of parameters. SINDy achieves MAPEs of $1.67\%$, $0.86\%$, and $0.93\%$, with 15, 19, and 20 active monomials, but the SINDy coefficients are unrestricted real numbers rather than nonnegative amplitudes, thus lacking physical interpretability of positive loss. SSI reduces these errors to \MapeA\%, \MapeB\%, and \MapeC\% while retaining an explicit expression with only 7--10 terms. Therefore, SSI provides the best balance among sparsity, explicit equation, and physical interpretability.

% \begin{figure}[t]
% \setlength{\abovecaptionskip}{3pt}
% \centering
% \includegraphics[width=\columnwidth]{method_comparison.png}
% \caption{Test MAPE of the empirical, numerical, and symbolic modeling methods for the three materials.}
% \label{fig:comparison}
% \end{figure}

% SINDy achieves MAPEs of $1.67\%$, $0.86\%$, and $0.93\%$, with 15, 19, and 20 active monomials. These errors are lower than SSI's, but the expressions retain more terms. More importantly for interpretation, the SINDy coefficients are unrestricted real numbers rather than nonnegative amplitudes. Its three coefficient sets include 8, 12, and 12 negative entries. They describe signed corrections in standardized logarithmic coordinates and cannot be directly interpreted as strengths of independent physical loss mechanisms. Recovering positive total loss by exponentiating~\eqref{eq:sindy} does not change this coefficient interpretation.

\section{Physical Implications and Limitations}
The discovered equations retain the analytical value of empirical loss models while allowing several nonlinear dependencies to be selected from data. Their power-law exponents, roll-off scales, and logarithmic corrections describe how the measured loss changes with frequency and flux density. Local logarithmic sensitivities can therefore be evaluated directly from the identified expression.

% For example, local logarithmic sensitivities can be evaluated directly from the identified expression:
% \begin{equation}
% \begin{aligned}
% \alpha_{\mathrm{eff}}&=\frac{\bar{f}}{\hat{\bar{P}}_v}
% \sum_{j\in\mathcal{S}}\xi_j\frac{\partial\theta_j}{\partial\bar{f}},\\
% \beta_{\mathrm{eff}}&=\frac{\bar{B}}{\hat{\bar{P}}_v}
% \sum_{j\in\mathcal{S}}\xi_j\frac{\partial\theta_j}{\partial\bar{B}}.
% \end{aligned}
% \label{eq:sensitivity}
% \end{equation}
% Unlike constant SE exponents, these sensitivities vary over the characterized domain. Together with $P_{\mathrm{core}}=V_e\hat{P}_v$ for effective core volume $V_e$ and consistent units, they support loss estimation and local magnetic-design calculations.

However, the retained functions only provide an effective macroscopic description rather than a unique microscopic loss separation. Several fixed-grid candidates are correlated, and more than one sparse combination may describe similar measured behavior. Specifically, the fixed library determines the available resolution, so a scaling law between grid values may require several nearby terms. A denser grid increases accuracy and correlation simultaneously, thus reducing sparsity.

Validation is limited to sinusoidal excitation at 25$^{\circ}$C within the measured envelopes. Temperature variation, DC bias, nonsinusoidal waveforms, and extrapolation require additional measurements and candidates.

\section{Conclusion}
This paper proposed a Symbolic Sparse Identification framework that discovers explicit magnetic core-loss equations from measured data. SSI casts core loss modeling as a data-driven symbolic regression problem without presuming a functional form, and constructs a comprehensively expanded physics-guided library of 103 symbolic candidate terms representing different loss dependencies. The dominant candidates are isolated from the library by a hybrid logarithmic-percentage objective with nonnegative AdamW updates and threshold pruning, and their coefficients are optimized to maximize accuracy. Experiments produce explicit equations with \TermsA, \TermsB, and \TermsC\ terms for FR78, FR79, and FR95, achieving test MAPEs of \MapeA\%, \MapeB\%, and \MapeC\%, respectively. The results present improved accuracy over the SE while maintaining analytical transparency. Comparisons with RF, FNN, and SINDy quantify the superior balance of SSI among accuracy, compactness, and physical interpretability. SSI thus provides a feasible approach to discover explicit equations for magnetic characterization and design. Future work will extend to broader temperature and excitation conditions.

\bibliographystyle{IEEEtran}
\bibliography{reference}

@IEEEtranBSTCTL{BSTcontrol,
  CTLdash_repeated_names = "no"
}

@misc{Wang2026Measurement,
  author={Wang, Haoyu and Hanson, Alex},
  title={Resonant Method-Based Fully Automated Core Loss Measurement System for Sub-{MHz} Magnetics With Switched Capacitor Sequence},
  year={2026},
  howpublished={arXiv:2605.30460},
  doi={10.48550/arXiv.2605.30460}
}

@article{7370781,
  author={Hanson, Alex J. and Belk, Julia A. and Lim, Seungbum and others},
  title={Measurements and Performance Factor Comparisons of Magnetic Materials at High Frequency},
  journal={IEEE Transactions on Power Electronics},
  year={2016}, volume={31}, number={11}, pages={7909--7925},
  doi={10.1109/TPEL.2015.2514084}
}

@article{Bertotti1988,
  author={Bertotti, Giorgio},
  title={General Properties of Power Losses in Soft Ferromagnetic Materials},
  journal={IEEE Transactions on Magnetics},
  year={1988}, volume={24}, number={1}, pages={621--630},
  doi={10.1109/20.43994}
}

@article{1457110,
  author={Steinmetz, Charles P.},
  title={On the Law of Hysteresis},
  journal={Proceedings of the IEEE},
  year={1984}, volume={72}, number={2}, pages={197--221},
  doi={10.1109/PROC.1984.12842}
}

@article{Reinert2001MSE,
  author={Reinert, J. and Brockmeyer, A. and De Doncker, Rik W. A. A.},
  title={Calculation of Losses in Ferro- and Ferrimagnetic Materials Based on the Modified Steinmetz Equation},
  journal={IEEE Transactions on Industry Applications},
  year={2001}, volume={37}, number={4}, pages={1055--1061},
  doi={10.1109/28.936396}
}

@inproceedings{Venkatachalam2002IGSE,
  author={Venkatachalam, Kapil and Sullivan, Charles R. and Abdallah, Tarek and others},
  title={Accurate Prediction of Ferrite Core Loss With Nonsinusoidal Waveforms Using Only Steinmetz Parameters},
  booktitle={2002 IEEE Workshop on Computers in Power Electronics},
  year={2002}, pages={36--41},
  doi={10.1109/CIPE.2002.1196712}
}

@article{5955126,
  author={Muhlethaler, Jonas and Biela, Juergen and Kolar, Johann W. and others},
  title={Improved Core-Loss Calculation for Magnetic Components Employed in Power Electronic Systems},
  journal={IEEE Transactions on Power Electronics},
  year={2012}, volume={27}, number={2}, pages={964--973},
  doi={10.1109/TPEL.2011.2162252}
}

@article{10169101,
  author={Serrano, Diego and Li, Haoran and Wang, Shukai and others},
  title={Why {MagNet}: Quantifying the Complexity of Modeling Power Magnetic Material Characteristics},
  journal={IEEE Transactions on Power Electronics},
  year={2023}, volume={38}, number={11}, pages={14292--14316},
  doi={10.1109/TPEL.2023.3291084}
}

@article{10502151,
  author={Li, Z. and Wang, L. and Liu, R. and others},
  title={A Data-Driven Model for Power Loss Estimation of Magnetic Materials Based on Multi-Objective Optimization and Transfer Learning},
  journal={IEEE Open Journal of Power Electronics},
  year={2024}, volume={5}, pages={605--617},
  doi={10.1109/OJPEL.2024.3389211}
}

@inproceedings{10509368,
  author={Sauter, Bailey and others},
  title={Integrating Equation-Based Methods with Random Forest Regression for Improved Accuracy of Magnetic Core Loss Modeling},
  booktitle={2024 IEEE Applied Power Electronics Conference and Exposition (APEC)},
  year={2024}, pages={410--415}
}

@article{1041954,
  author={Goodenough, J. B.},
  title={Summary of Losses in Magnetic Materials},
  journal={IEEE Transactions on Magnetics},
  year={2002}, volume={38}, number={5}, pages={3398--3408},
  doi={10.1109/TMAG.2002.802741}
}

@article{Brunton2016SINDy,
  author={Brunton, Steven L. and Proctor, Joshua L. and Kutz, J. Nathan},
  title={Discovering Governing Equations From Data by Sparse Identification of Nonlinear Dynamical Systems},
  journal={Proceedings of the National Academy of Sciences},
  year={2016}, volume={113}, number={15}, pages={3932--3937},
  doi={10.1073/pnas.1517384113}
}

@inproceedings{Kingma2015Adam,
  author={Kingma, Diederik P. and Ba, Jimmy},
  title={{Adam}: A Method for Stochastic Optimization},
  booktitle={International Conference on Learning Representations (ICLR)},
  year={2015},
  url={https://arxiv.org/abs/1412.6980}
}

@inproceedings{Loshchilov2019AdamW,
  author={Loshchilov, Ilya and Hutter, Frank},
  title={Decoupled Weight Decay Regularization},
  booktitle={International Conference on Learning Representations (ICLR)},
  year={2019},
  url={https://arxiv.org/abs/1711.05101}
}

@INPROCEEDINGS{10977470,
  author={Wang, Haoyu and Zhang, Yuanxin and Mou, Di and others},
  booktitle={2025 IEEE Applied Power Electronics Conference and Exposition (APEC)}, 
  title={High-Frequency Conditioning Circuits for Power-Related Information Extraction in Non-Sinusoidal Power Electronic Systems}, 
  year={2025},
  volume={},
  number={},
  pages={2857-2864},
  doi={10.1109/APEC48143.2025.10977470}
}

@INPROCEEDINGS{11121265,
  author={Wang, Haoyu and Brown, Alyssa and Wu, Yihao and others},
  booktitle={2025 IEEE 26th Workshop on Control and Modeling for Power Electronics (COMPEL)}, 
  title={Automatic Loss Measurement System for MHz Magnetics Using High-Frequency Conditioning Circuits}, 
  year={2025},
  volume={},
  number={},
  pages={1-7},
}

@ARTICLE{9893537,
  author={Wang, Haoyu and Zeng, Yangbin and Ji, Shiqi and others},
  journal={IEEE Transactions on Industrial Electronics}, 
  title={ZVS Soft Switching Operation Region Analysis of Modular Multi Active Bridge Converter Under Single Phase Shift Control}, 
  year={2023},
  volume={70},
  number={7},
  pages={6865-6875},
  doi={10.1109/TIE.2022.3204954}
}

@ARTICLE{10242261,
  author={Wang, Haoyu and Mou, Di and Ji, Shiqi and others},
  journal={IEEE Transactions on Industrial Electronics}, 
  title={Universal Phase-Shift Modulation Scheme and Efficiency Optimization for Modular Multiactive Bridge Converter}, 
  year={2024},
  volume={71},
  number={7},
  pages={7312-7321},
  doi={10.1109/TIE.2023.3306403}
}

@ARTICLE{10132875,
  author={Wang, Haoyu and Ji, Shiqi and Mou, Di and others},
  journal={IEEE Journal of Emerging and Selected Topics in Power Electronics}, 
  title={Switching Characterization and Power Loss Optimization for Modular Multiactive Bridge Converter Under Common Phase Shift Control}, 
  year={2023},
  volume={11},
  number={4},
  pages={3924-3936},
  doi={10.1109/JESTPE.2023.3279658}
}

@ARTICLE{10319663,
  author={Wang, Haoyu and Ji, Shiqi and Mou, Di and others},
  journal={IEEE Journal of Emerging and Selected Topics in Industrial Electronics}, 
  title={High-Frequency-Link-Based Reactive-Power Optimal Control for Modular Multi-Active Bridge Converter}, 
  year={2024},
  volume={5},
  number={3},
  pages={1333-1337},
  doi={10.1109/JESTIE.2023.3333311}
}

@INPROCEEDINGS{10362302,
  author={Wang, Haoyu and Mou, Di and Ji, Shiqi and others},
  booktitle={2023 IEEE Energy Conversion Congress and Exposition (ECCE)}, 
  title={Comparison and Improvement of ZVS Operation Under Different Modulation Strategies for Modular Multi Active Bridge Converters}, 
  year={2023},
  volume={},
  number={},
  pages={2556-2563},
  doi={10.1109/ECCE53617.2023.10362302}
}

@INPROCEEDINGS{9628583,
  author={Xiang, Minjiang and Liu, Yang and Wang, Haoyu and others},
  booktitle={2021 IEEE 1st International Power Electronics and Application Symposium (PEAS)}, 
  title={Design and Analysis of a 75 kVA High-Frequency-Link Based Three-Port Power Electronic Transformer}, 
  year={2021},
  volume={},
  number={},
  pages={1-7},
  doi={10.1109/PEAS53589.2021.9628583}
}

@ARTICLE{10194423,
  author={Mou, Di and Wang, Haoyu and Yuan, Liqiang and others},
  journal={IEEE Transactions on Industrial Electronics}, 
  title={High-Efficiency Time-Division Multiplexing Modulation Technology for Modular Multiactive Bridge Converters}, 
  year={2024},
  volume={71},
  number={6},
  pages={5745-5754},
  doi={10.1109/TIE.2023.3292848}
}

@ARTICLE{11358392,
  author={Zheng, Jialin and Wang, Haoyu and Zeng, Yangbin and others},
  journal={IEEE Transactions on Industrial Electronics}, 
  title={Physics-Embedded Neural ODEs for Sim-to-Real Edge Digital Twins of Hybrid Power Electronics Systems}, 
  year={2026},
  volume={73},
  number={6},
  pages={8616-8627},
  doi={10.1109/TIE.2025.3645414}
}

@ARTICLE{11366028,
  author={Zheng, Jialin and Wang, Haoyu and Zeng, Yangbin and others},
  journal={IEEE Transactions on Industrial Electronics}, 
  title={Neural Surrogate Solver for Efficient Edge Inference of Power Electronic Hybrid Dynamics}, 
  year={2026},
  volume={73},
  number={6},
  pages={9523-9528},
  doi={10.1109/TIE.2025.3642411}
}
\end{document}